\documentclass[11pt]{article}

\usepackage{dialogue}

\usepackage{ifxetex}
\ifxetex
    \usepackage{fontspec}
    \setromanfont{Times New Roman}
\else
  \usepackage{cmap}
  \usepackage[T2A,T1]{fontenc}
  \usepackage[utf8]{inputenc}
  \usepackage{times}
  \usepackage{latexsym}
  \usepackage{substitutefont}
  \substitutefont{T2A}{\familydefault}{cmr}
\fi

\usepackage[russian,british]{babel}
\usepackage{url}
\usepackage{pgf}
\usepackage{adjustbox}

\usepackage{covington} 

\usepackage{hyperref}

\dialogfinalcopy 

\title{RuOpinionNE-2024: Extraction of Opinion Tuples \\from Russian News Texts}

\author{ 
    Natalia Loukachevitch \\
    Lomonosov Moscow  \\
    State University  \\
    {\tt louk\_nat@mail.ru} \\ 
    \And
    Natalia Tkachenko \\
    Lomonosov Moscow  \\
    State University  \\
    {\tt nataliya.m.tkachenko@gmail.com} \\  
    \AND
    Anna Lapanitsyna \\
    Lomonosov Moscow  \\
    State University  \\
    {\tt anna.lapachka@gmail.com} \\ 
    \And
    Mikhail Tikhomirov \\
    Lomonosov Moscow  \\
    State University  \\
    {\tt tikhomirov.mm@gmail.com} \\  
    \And
    Nicolay Rusnachenko \\
    Bauman Moscow State \\
    Technical University  \\
    {\tt rusnicolay@gmail.com} \\ 
}

\date{}

\begin{document}
\maketitle
\begin{abstract}
In this paper, we introduce the Dialogue Evaluation shared task on extraction of structured opinions from Russian news texts. The task of the contest is to extract opinion tuples for a given sentence; the tuples are composed of a sentiment holder, its target, an expression and sentiment from the holder to the target. In total, the task received more than 100 submissions. The participants experimented mainly with large language models in zero-shot, few-shot and fine-tuning formats. The best result on the test set was obtained with fine-tuning of a large language model. We also compared 30 prompts and 11 open source language models with 3-32 billion parameters in the 1-shot and 10-shot settings and found the best models and prompts.

  
  \textbf{DOI:} 10.28995/2075-7182-2022-20-XX-XX
\end{abstract}

\selectlanguage{russian}
\begin{center}
  \russiantitle{RuOpinionNE-2024:  извлечение кортежей мнений \\из новостных текстов на русском языке}

  \medskip \setlength\tabcolsep{0.3cm}

\begin{table}[!htp]
    \begin{tabular}{cccccc}
    \multicolumn{3}{c}{\hspace{2cm}{\bf Лукашевич Н.В.}}                                                   & \multicolumn{3}{c}{{\bf Ткаченко Н.М.}}                                \\
    \multicolumn{3}{c}{\hspace{2cm}МГУ им. Ломоносова}                                               & \multicolumn{3}{c}{МГУ им. Ломоносова}                           \\
    \multicolumn{3}{c}{\hspace{2cm}Москва, Россия}                            & \multicolumn{3}{c}{Москва, Россия}        \\
    \multicolumn{3}{c}{\hspace{2cm} louk\_nat@mail.ru}                                                & \multicolumn{3}{c}{nataliya.m.tkahenko@gmail.com}                \\
    \\
    \multicolumn{2}{c}{{\bf Лапаницына А.М.}}                             & \multicolumn{2}{c}{{\bf Тихомиров М.M.}}            & \multicolumn{2}{c}{{\bf Русначенко Н.Л.}}       \\
    \multicolumn{2}{c}{МГУ им. Ломоносова} & \multicolumn{2}{c}{МГУ им. Ломоносова}      & \multicolumn{2}{c}{МГТУ им. Н.Э. Баумана} \\
    \multicolumn{2}{c}{Москва, Россия}     & \multicolumn{2}{c}{Москва, Россия}          & \multicolumn{2}{c}{Москва, Россия}        \\
    \multicolumn{2}{c}{anna.lapachka@gmail.com}                   & \multicolumn{2}{c}{tikhomirov.mm@gmail.com} & \multicolumn{2}{c}{rusnicolay@gmail.com} 
    \end{tabular}
\end{table}
\end{center}

\begin{abstract}
 В этой статье описано новое тестирование по извлечению  мнений из русскоязычных новостных текстов, организованное в рамках серии тестирований Dialogue Evaluation. Задача тестирования состоит в извлечении кортежей мнений для заданного предложения; кортежи состоят из источника мнения, объекта мнения, оценочного выражения и тональности источника по отношению к объекту. Всего на тестирование было подано  более 100 прогонов. Участники экспериментировали в основном с большими языковыми моделями в форматах zero-shot, few-shot и fine-tuning. Лучший результат на тестовом наборе был получен на основе дообучения (fine-tuning) большой языковой модели. Также было проведено  сравнение 30 промптов  и 11 языковых моделей с открытым исходным кодом размером 3-32 миллиардов параметров в 1-shot и 10-shot режимах и выявлены лучшие модели и промпты.
 
  \textbf{Ключевые слова:} таргетированный анализ тональности, именованная сущность, новостные тексты 
\end{abstract}
\selectlanguage{british}

\section{Introduction}
\label{intro}
Sentiment analysis is one of the most actively developing areas in natural language processing. At the beginning of studies in this area, the task was to extract the overall sentiment of a user review or message in a social network. Currently, there is a wide variety of problem statements, including the identification of sentiment toward a specific entity, an aspect (part or characteristic) of this entity, and a related task of extracting a position (stance) on a certain discussion topic.  Modern methods allow us to get closer to a more complete opinion recognition \cite{liu2012sentiment}, extracting opinion tuples,  which include a source of the opinion, an object of the opinion, sentiment, and other opinion components \cite{barnes2022semeval}.

\begin{figure}
    \centering
    \includegraphics[width=0.407\linewidth]{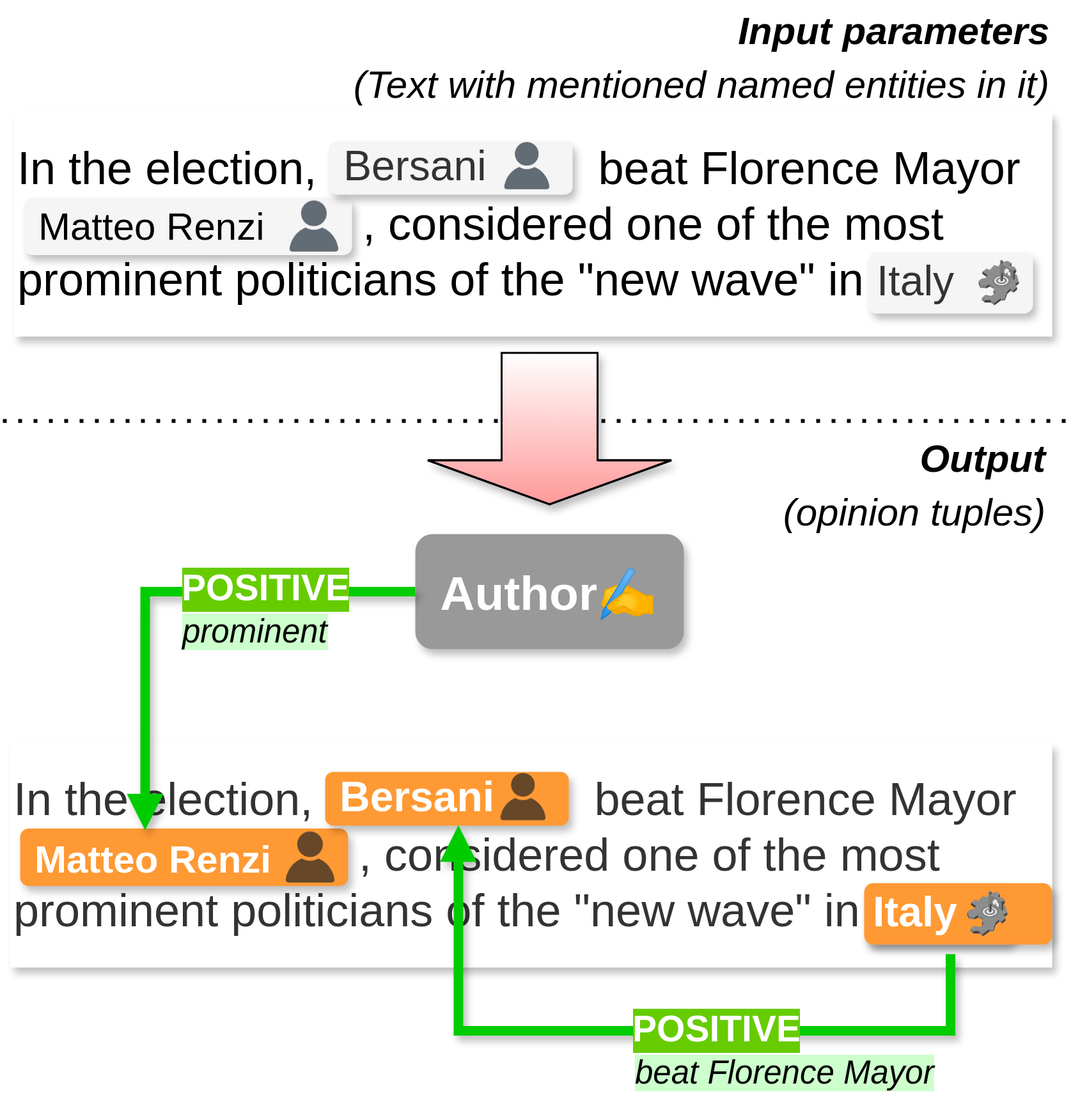}
    \includegraphics[width=0.49\linewidth]{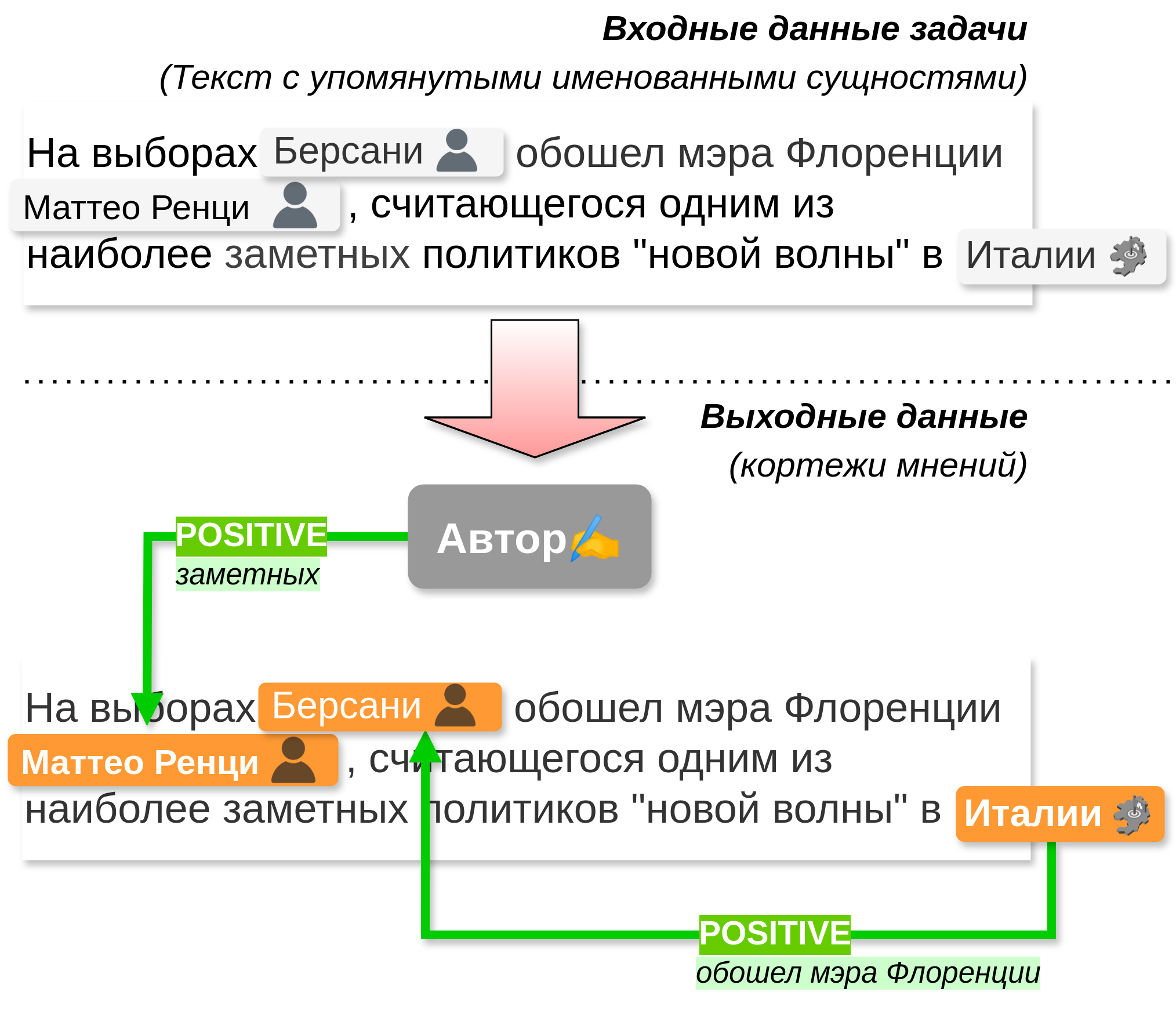}
    \caption{Example of opinion extraction with explanations according to the RuOpinionNE-2024 task. Two opinion tuples are shown: (Italy, Bersany, beat Florence Mayor, positive) and (AUTHOR, Matteo Renzi, prominent, positive)}
    \label{fig:task-example}
\end{figure}

In this paper, we present the RuOpinionNE shared task on extraction of opinions from Russian news texts organized in the framework of the Dialogue evaluation competitions. The aim of the task is to extract opinion tuples for a given sentence; the tuples are composed of a sentiment holder, its target, an expression and sentiment from the holder to the target. News texts are a very interesting source for extracting opinions because such texts contain numerous entities, which can be the holder or the target of opinion or both. In the same sentence,  positive and negative attitudes can be met.  At the same time, most entities are mentioned in neutral contexts. The extraction of such tuples from news texts allows a better understanding of attitudes between different subjects discussed in the current flow of news.
Figure~\ref{fig:task-example} depicts an example of opinion tuples extracted from a Russian text (right) along with the translated version of the related example (left). In the example, the author is positive to Matteo Renzi, because he is mentioned as a prominent politician. In addition, Italy is positive to Bersani because Italians voted for him.

\section{Related Work}
The extraction of opinions from texts should aim to identify tuples that comprise several components such as the source of the opinion, the target, maybe the aspect (part, characteristics) of the target and the sentiment \cite{liu2012sentiment}.  Currently, in numerous sentiment-oriented works, the ultimate task is often reduced to simpler subtasks such as general sentiment of a text (sentence), target-oriented sentiment (the identification of the opinion source is not included in the task), and others \cite{barnes-etal-2021-structured}.

Barnes et al. \cite{barnes-etal-2021-structured} suggest extracting opinions as tuples $(h, t, e, p)$ where $h$ is a holder who expresses a polarity $p$ towards a target $t$ through a sentiment expression $e$. Extracted tuples can be represented as a graph comprising a set of labeled
nodes and a set of unlabeled edges connecting pairs
of nodes. 
Empty holders and targets are allowed.  The authors used a reimplementation of the neural syntactic parser by Dozat and Manning \cite{dozat2022deep} to extract such tuples from texts.   The base of the network model is a bidirectional LSTM~\cite{hochreiter1997long,schuster1997bidirectional} (BiLSTM), which creates contextualized representations $[c_1, \ldots , c_n] = BiLSTM(w_1, . . . , w_n)$, where $w_i$ is the concatenation of a word embedding, POS tag embedding, lemma embedding, character embedding created by a character-based LSTM for the $i$'th token, and also BERT~\cite{devlin2018bert}  contextualized embeddings. The obtained embeddings are then processed by two feedforward neural networks (FNN),
creating specialized representations for potential
heads and dependents as $h_i = FNNhead(c_i)$ and $d_i = FNNdep(c_i)$ respectively. The scores for each possible relation between found items are computed by a final bilinear
transformation.

In 2022,  the multilingual contest on the extraction of opinion tuples was organized at the SemEval 2022 conference \cite{barnes2022semeval}. The evaluation collection comprises mainly review datasets and a single news collection MPQA. The organizers provided two baselines in the contest: 1) a dependency graph prediction model, and 2) a
sequence-labeling pipeline based on three separate BiLSTM models to extract
holders, targets, and expressions. The outputs of the models were used to train a relation prediction model that determines sentiment relations between found entities. The best results in the competition were achieved by applying a modified version of the above mentioned sentiment graph analysis model with contextualized embeddings obtained with RoBERTa\textsubscript{Large} \cite{liu2019robertarobustlyoptimizedbert}.

In aspect-oriented sentiment analysis, 
several tuple configurations are used to structure the content.
Such configurations are: Aspect-Sentiment Triplet Extraction (ASTE), which should produce \{aspect,
opinion, sentiment category\} triplets, such as “⟨menu, great, positive⟩”; and Aspect-Sentiment Quadruplet Extraction/Prediction (ASQE/ASQP) that outputs \{aspect, opinion, aspect category, sentiment category\} quadruplets, such as “⟨menu, great, general, positive⟩” \cite{hua2024systematic}.

The authors of \cite{zhang2024sentiment} tested zero-shot large language models (LLM) in various sentiment analysis tasks, including aspect tuple extraction: Aspect-Sentiment Triplet Extraction (ASTE) and Aspect-Sentiment Quadruplet Extraction/Prediction (ASQE/ASQP). They found that zero-shot prompting of larger LLMs has very low results in these tasks, fine-tuned T5\textsubscript{large} model~\cite{raffel2020exploring} (small LLM) achieved much higher results:  about 24 percent points for ASTE on average and 30 percent points for ASQE/ASQP.


For Russian, there were some studies on targeted sentiment analysis, such as aspect-based sentiment analysis \cite{loukachevitch2015sentirueval,chumakov2023generative} or entity-oriented sentiment analysis \cite{loukachevitch2015entity,golubev2020improving,pronoza2021detecting,golubev2023rusentne}. In 2023, the RuSentNE competition devoted to extraction of sentiment tovards named entities in news texts was organized.  The best approaches on the  RuSentNE  evaluation  were BERT-based ensembles \cite{devlin2018bert,golubev2023rusentne}. Currently, the best result in this formulation of the problem has been achieved with fine-tuning of the Flan-T5 model \cite{rusnachenko2024large}.
In \cite {rusnachenko2019distant}, the authors studied the extraction of triples \{holder, target, sentiment\} at the document level in the RuSentRel dataset, which comprised 73 analytical texts on international relations.   The current work is the first study on extracting sentiment tuples at the sentence level for the Russian language.

\section{RuSentNE Corpus}
The RuOpinionNE-2024 dataset is based on the annotated RuSentNE corpus, which includes news texts written in Russian. In the first stage of the annotation, the RuSentNE texts were annotated with named entities, including such types as PERSON, ORGANIZATION, PROFESSION, COUNTRY, CITY, NATIONALITY and so on. In the next stage, positive or negative relations between entities (including the author's position towards mentioned entities) were established. Finally, for each annotated attitude, the evidence (that is, an expression that serves as a key element for extracting the opinion) was carefully annotated and checked (Figure \ref{fig:examples}). The annotations allowed us to create a dataset, in which sets of opinion tuples (holder, target, sentiment, expression) are assigned to each text fragment.

The source of the opinion (Opinion holder) can be:
(i)~the author of the text, 
(ii)~the author of the quote, or 
(iii)~another mentioned entity. 
The opinion can be \textit{implicit}, i.e. expressed through the actions of the entity. 
For example, in the sentence “X fired Y”, it is assumed that there is an implicit negative opinion of X towards Y.



The first version of the created corpus became a basis for the RuSentNE-2023 open evaluation, which required the recognition of sentiment in relation to a given named entity \cite{golubev2023rusentne}.
The task was to predict whether the content of a given sentence is positive or negative towards a given entity \cite{golubev2023rusentne}. 


\section{RuOpinionNE-2024 Dataset and Task Description}
\label{sec:dataset}

\begin{figure}
    \centering
    \includegraphics[width=0.9\linewidth]{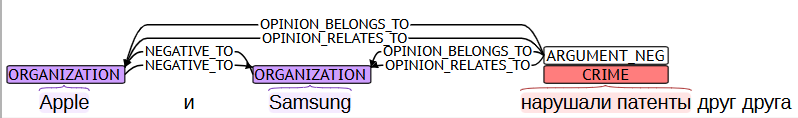}
    \caption{Example of opinion annotation for the RuOpinionNE-2024 task}
    \label{fig:examples}
\end{figure}

The task of the RuOpinionNE-2024 evaluation is as follows: for given fragments of news articles (mainly sentences), it is necessary to extract all tuples:
$(H, T, P, E)$,
where $H$ is an opinion holder who expresses a polarity $P$ towards a target $T$ through a sentiment expression $E$. 
There can be sentences without any tuple.
Holders and Targets can be entities of the following types: PERSON, ORGANIZATION, COUNTRY, CITY, REGION, PROFESSION, NATIONALITY, IDEOLOGY.
Pronouns are not included in tuples.
Holders can also be empty (NULL) for general opinion or AUTHOR for the author's opinion. Targets and expressions are never empty.
Fragmented entities are possible. The polarity represents sentiment label which could be positive or negative. Table \ref{tab:Examples} shows examples of tuples extracted from the example sentence (Figure \ref{fig:examples}).

\begin{table}[!htp]
    \centering
    \begin{tabular}{lccl}
    \hline
   Holder         &Target& Polarity &  Expression\\
    \hline
    
   Apple& Samsung &  NEG & violated patents\\
   Samsung & Apple &  NEG & violated patents\\
    \hline
    \end{tabular}
    \caption{Examples of sentiment opinions in RuOpinionNE-2024 dataset (Figure \ref{fig:examples}) }
    \label{tab:Examples}
\end{table}

Table \ref{tab:distribution} contains the distribution of opinion tuples in train, validation and test sets of the RuOpinionNE-2024 dataset. It can be seen that about 40\% of sentences do not contain opinion tuples. Sentences with opinions contain two opinion tuples on average. In about 20\% of annotated tuples, the holder is absent, and about 10\% of tuples are annotated as the author's opinion. 

\begin{table}[!htp]
    \centering
    \begin{tabular}{llccccc}
    \hline
    Type & Stage & \#Sent. & \#Sent. & \#Tuples   &\#Tuples&\#Tuples with  \\
      &   &        &w/o sentiment&     &w/o holder&author as holder\\
    \hline
    train      & Train       &  2556    & 1062 & 2904 & 484&173\\
    validation & Development &  1316    & 547 & 1612 & 436&196\\
    test       & Final       &  804    & 216 & 1172 & 334&158\\
    \hline
    Total & &                   4676   &1825&5688&1254&527\\
    \hline
    \end{tabular}
    \caption{Distribution of  sentiment scores in training, validation and test sets.}
    \label{tab:distribution}
\end{table}

The data was prepared in the JSON lines format\footnote{\url{https://jsonlines.org/}}, where a line corresponds to a sentence with found opinion tuples or with an empty set of opinion tuples.

The main metric for the task is the F1 measure.  When evaluating recall, each reference tuple was compared with the entire list of predicted tuples for the corresponding text fragment. The reference tuple was considered found if among the predictions at least one tuple detected, for which:
\begin{itemize}
\item 1) holder, target, expression intersect with the reference holder, target, expression by at least one token, respectively (it is necessary that each component is at least partially detected and all of them are combined into a single tuple); 
\item 2) the polarity of the same tuple coincides with the polarity of the reference.
\end{itemize}

The cases of holder = AUTHOR or NULL were processed as special tokens that required complete match.   For example,  for holder overlap:
\begin{itemize}
\item holder\_overlap = (number of common tokens of the reference and the prediction)/(number of reference tokens).
\end{itemize}

Target overlap and expression overlap were calculated in a similar way. If there were several options, the best tuple was selected. 
If the tuple was found, a weighted score was calculated for it, which reflected the coverage of the reference tuple by the predicted one, 
averaged over the three components. 
Weighted score and recall were calculated as follows:
\begin{itemize}
\item weighted\_score = (holder\_overlap + target\_overlap + expression\_overlap)/3, if each overlap >0 and the polarity matches, and 0 otherwise;
\item recall = (sum of weighted\_score over all reference tuples)/(number of reference tuples).
\end{itemize}

Precision was calculated in the same way, but for it, the coverage of the set of predicted tuples by the set of reference tuples was estimated.  Both metrics were calculated not for sentences separately, but for the entire dataset, and then combined into the F1 metric.


\section{Results of Evaluation}
\label{sec:results}

 As a baseline (baseline\_model), we adopt Qwen2.5-32B-Instruct model~\cite{qwen2025qwen25technicalreport} (32 billion parameters) in the few-shot format (10 examples). The complete textual prompt utilized for inferring results is shown in Appendix~\ref{sec:baseline}.
 
In our competition,  we received more than 100 submissions.  Table \ref{tab:dev} shows the results of the models at the development stage, table \ref{tab:final} presents the results of the models at the final stage of the competition.
In the following, we describe the approaches of participants.
\begin{table}[!htp]
    \centering
    \begin{tabular}{lc}
    \hline
    Participant & F1 \\
    \hline
    Zholtikov\_Michail & 0.34    \\
    msuai              & 0.28    \\
    iarnv              & 0.21    \\
    baseline\_model    & 0.17    \\
    \hline
    \end{tabular}
    \caption{Results of the Development evaluation stage of RuOpinionNE-2024 }
    \label{tab:dev}
\end{table}

\begin{table}[h]
    \centering
    \begin{tabular}{lcc}
    \hline
    Participant        & F1 \\
    \hline
    \textbf{VatolinAlexey}     & 0.41  \\
    \textbf{RefalMachine}      & 0.35  \\
    \textbf{msuai}             & 0.33  \\
    iarnv              & 0.28  \\
    Zholtikov\_Michail & 0.24  \\
    baseline\_model    & 0.24  \\
    vitalymegabyte     & 0.20  \\
    utmn               & 0.11  \\
    \hline
    \end{tabular}
    \caption{Results of the Final evaluation stage of RuOpinionNE-2024}
    \label{tab:final}
\end{table}

\textbf{VatolinAlexey.} 
The participant experimented with unsupervised and supervised approaches.  In an unsupervised approach, the LlaMA-3.3-70B\footnote{\url{https://huggingface.co/meta-llama/Llama-3.3-70B-Instruct}}~\cite{grattafiori2024llama3herdmodels} model was used. The model could not calculate exact indices of the items for opinion tuples in the substring, therefore, the final prompt did not contain the task to determine the indices. Instead, the algorithm was written for fuzzy search of a substring in a given sentence.  The best result of the unsupervised experiments was F1 = 0.22. 

As a supervised approach, the QLora~\cite{dettmers2023qlora} adapter for the LLaMA-3.3-70B model with FP4 precision was trained on the training data.  Only text was fed as input without any prompts.  The output was similar to what is described in the task, i.e. a JSON with a list of opinions.  The model could produce different results after each run due to randomization,  so the model was run several times and the results were aggregated. Several aggregation strategies were tested: 
(1) most frequent answer (\texttt{most\_common}), 
(2) $N$ most frequent answers (\texttt{most\_common\_n}), 
(3) combining all answers (\texttt{combine}). 
The best result on the test data was obtained with $N = 5$, strategy \texttt{most\_common\_n} with the result F1~=~0.41.

\textbf{RefalMachine.} 
The participant experimented with zero-shot, few-shot~\cite{brown2020languagemodelsfewshotlearners} and fine-tuning approaches. 
The best results were achieved by fine-tuning the Qwen2.5-32B-Instruct model and a prompt in the instruction format, with an F1 score of 0.35 at the Final evaluation stage.

\textbf{msuai.} 
The participant experimented with unsupervised and supervised formats of the LLM application:
Mistral Large 2\footnote{\url{https://huggingface.co/mistralai/Mistral-Large-Instruct-2407}},  GPT4-o\cite{openai2024gpt4ocard}.
Models were used with prompts containing 12-15 examples. The examples were chosen according to semantic similarity with a target sentence calculated using  
BERT\textsubscript{large-uncased} model for sentence embeddings in Russian language\footnote{\url{https://huggingface.co/ai-forever/sbert_large_nlu_ru}} or 
Sentence RuBERT\footnote{\url{https://huggingface.co/DeepPavlov/rubert-base-cased-sentence}}
models. 
This approach achieved the highest participant F1 score of 0.33 at the Final evaluation stage.

\textbf{Zholtikov\_Michail}. The participant experimented with the instruction tuned versions of 
Mistral-7B-v0.3\footnote{\url{https://huggingface.co/mistralai/Mistral-7B-Instruct-v0.3}}, 
MetaLlama-3-8B\footnote{\url{https://huggingface.co/meta-llama/Meta-Llama-3-8B-Instruct}} and 
Vikhr-7B-0.3\footnote{\url{https://huggingface.co/Vikhrmodels/Vikhr-7B-instruct_0.3}}. 
The format of experiments: zero-shot and supervised learning. 
The best results F1=0.34 were obtained by instruction tuned version of Mistral-7B-v0.3 at the Development stage, and F1 = 0.24 at the Final evaluation stage.

\textbf{vitalymegabyte}. The participant applied traditional NER, splitting polar expressions to separate tags: Polar\_expression\_POS and Polar\_expression\_NEG. To extract relationships between extracted items, self-attention was used to build the relationship matrix between items.

Thus, we can see that almost all participants experimented with generative models in zero-shot, few-shot or fine-tuning regimes. The best results were achieved by parameter-efficient fine-tuning of a quite large language model LlaMA-3.3-70B-Instruct with 70 billion parameters.

Compared with the results of the Semeval 2022 competition \cite{barnes-etal-2021-structured}, it could be noted that the Semeval 2022 results for the review datasets are much higher, achieving 0.76 F-measure on an English review dataset, but the best results on the MPQA news dataset is approximately the same: about 0.44 F-measure.

\section{Experiments with RuOpinionNE Data in the Framework of Student Course}
RuOpinionNE data were used as experimental data in the student course "Practical aspects of LLM training", which was taught by Mikhail Tikhomirov at the Faculty of Computational Mathematics and Cybernetics of Lomonosov Moscow State University.

The students constructed 30 prompts and experimented with them in zero-shot or few-shot regimes on the RuOpinionNE data. The collected prompts were fed into eleven language models in 1-shot or 10-shot formats. The following language models were used in experiments:
\begin{itemize}
    \item LLaMA-3.1-8B-Instruct \footnote{\url{https://huggingface.co/meta-llama/Meta-Llama-3-8B-Instruct}}, LLaMA-3-8B-Instruct \footnote{\url{https://huggingface.co/meta-llama/Llama-3.1-8B-Instruct}};
    \item Mistral-Nemo-Instruct-2407 (12.2B parameters) \footnote{\url{https://huggingface.co/mistralai/Mistral-Nemo-Instruct-2407}};
\item Qwen2.5-32B-Instruct, Qwen2.5-14B-Instruct,  Qwen2.5-7B-Instruct, Qwen2.5-3B-Instruct \footnote{\url{https://huggingface.co/collections/Qwen/qwen25-66e81a666513e518adb90d9e}};
\item T-lite-it-1.0 (7.6B parameters)\footnote{\url{https://huggingface.co/t-tech/T-lite-it-1.0}}; 
\item OpenChat-3.5-0106  (7B parameters) \footnote{\url{https://huggingface.co/openchat/openchat-3.5-0106}};
\item RuAdapt-LLaMA3 (8B parameters) \footnote{\url{https://huggingface.co/RefalMachine/ruadapt\_llama3\_8b\_instruct\_extended\_lep\_ft}}~\cite{tikhomirov2024facilitating};
\item Saiga-LLaMA3-8B (8B parameters) \footnote{\url{https://huggingface.co/IlyaGusev/saiga\_llama3\_8b}}
\end{itemize}

\begin{table}[h]
\centering
\begin{tabular}{lcc}
\hline 
\textbf{model\_name} & \textbf{k=10} & \textbf{k=1} \\
\hline
Qwen2.5-32B-Instruct & \textbf{0.195} & \textbf{0.158} \\
Mistral-Nemo-Instruct-2407 & \underline{0.190} & 0.112 \\
Qwen2.5-7B-Instruct & 0.184 &  \underline{0.139} \\
Saiga-LLaMA3-8B & 0.179 & 0.091 \\
T-lite-it-1.0 & 0.157 & 0.096 \\
LLaMA-3-8b-Instruct & 0.153 & 0.119 \\
Qwen2.5-14B-Instruct & 0.145 & 0.121 \\
Meta-LlaMA-3.1-8B-Instruct & 0.141 & 0.090 \\
RuAdapt-LLaMA3 & 0.123 & 0.073 \\
OpenChat-3.5-0106 & 0.113 & 0.087 \\
Qwen2.5-3B-Instruct & 0.091 & 0.088 \\
\hline
\end{tabular}
\caption{Average model quality in 10-shot and 1-shot settings for the RuOpinionNE test set. The best results are in bold, the second best results are underlined}
\label{tab:models_comparison_average}
\end{table}

\begin{table}[h]
\centering
\begin{tabular}{lcc}
\hline 
\textbf{model\_name} & \textbf{k=10} & \textbf{k=1} \\
\hline
Qwen2.5-32B-Instruct & \textbf{0.229} & \textbf{0.204} \\
Mistral-Nemo-Instruct-2407 & \underline{0.211} & \underline{0.157} \\
Qwen2.5-7B-Instruct & 0.199 & 0.168 \\
Saiga-LLaMA3-8B & 0.193 & 0.118 \\
LLaMA-3.1-8B-Instruct & 0.173 & 0.110 \\
T-lite-it-1.0 & 0.171 & 0.119 \\
LLaMA-3-8B-Instruct & 0.169 & 0.154 \\
Qwen2.5-14B-Instruct & 0.169 & 0.144 \\
RuAdapt-LLaMA3 & 0.134 & 0.104 \\
OpenChat-3.5-0106 & 0.132 & 0.108 \\
Qwen2.5-3B-Instruct & 0.120 & 0.119 \\
\hline 
\end{tabular}
\caption{Maximum model quality in 10-shot and 1-shot settings for the RuOpinionNE test set. The best results are in bold, the second best results are underlined}
\label{tab:models_comparison_max}
\end{table}

Thus, the set of models in experiments comprises a larger model Qwen2.5-32B-Instruct (32 billion parameters), average-sized models with 12-13 billion parameters (Qwen2.5-14B-Instruct,  Mistral-Nemo-Instruct-2407),  smaller models with 7-8 billion parameters, and the smallest model with 3 billion parameters (Qwen2.5-3B-Instruct). The models in experiments include three models adapted to the Russian language with 7-8 billion parameters: T-lite-it-1.0, Saiga-LLaMA3-8B and RuAdapt-LLaMA3. 

Table \ref{tab:models_comparison_average} shows the performance of each model averaged on all prompts. It can be seen that the averaged best results are obtained on 10-shot prompts by larger models  Qwen2.5-32B-Instruct and  Mistral-Nemo-Instruct-2407; the averaged best results on 1-shot prompts are achieved by Qwen2.5-32B-Instruct. The   Qwen2.5-7B-Instruct  model obtained the best averaged results among all the smaller models.

Among the Russian-oriented models, the best results for the 10-shot setting were achieved by Saiga-LLaMA3-8B. For 1-shot prompts,  similar results were obtained by T-lite-it-1.0 and Saiga-LLaMA3-8B. In both cases, the results of the Russian models are worse than the results of the similar-sized model Qwen2.5-7B-Instruct.

Table \ref{tab:models_comparison_max} shows the best results for each model. The best results are obtained by larger models:  Qwen2.5-32B-Instruct and  Mistral-Nemo-Instruct-2407. Among the smaller models, the Qwen2.5-7B-Instruct model achieved the highest results. Among Russian models, the Saiga-LLaMA3-8B model obtained better results in the 10-shot setting and similar results with T-lite-it-1.0 in the 1-shot setting.

Table \ref{tab:instructions_comparison} presents the results of all instructions averaged on the models. It can be seen that the obtained results are very close to each other in average quality. This means that there does not exist a single prompt, which is best for all or most models.

\begin{table}[h]
\centering
\begin{tabular}{c c c}
\hline 
\textbf{instruction\_id} & \textbf{k=10} & \textbf{k=1} \\
\hline
28 & 0.162 & 0.113 \\
19 & 0.159 & 0.112 \\
17 & 0.158 & 0.120 \\
5 & 0.157 & 0.103 \\
0 & 0.157 & 0.105 \\
7 & 0.156 & 0.099 \\
23 & 0.155 & 0.108 \\
13 & 0.154 & 0.081 \\
12 & 0.154 & 0.112 \\
16 & 0.154 & 0.118 \\
\hline 
\end{tabular}
\caption{Results of all instructions averaged on the models in 10-shot and 1-shot settings for the RuOpinionNE test set. }
\label{tab:instructions_comparison}
\end{table}

\section{Analysis of Errors}

In this section,  we provide an analysis of the submitted results. 
To explore the cases that arise from disagreement between gold standard annotation and tuples automatically extracted  by participating models, we use two different modes for analysis:
\begin{itemize}
    \item \textbf{Extracted opinions mentioned in  manual annotation}: absence of annotated tuples in the submitted results.
    \item \textbf{Extracted opinions not mentioned in the gold standard}: analysis of tuples that were found in the submitted results but were not indicated in the gold annotation. 
\end{itemize}

We consider examples from the final stage of the competition.
We limit the scope of methods by overviewing LLM-based submissions, according to the Section~\ref{sec:results}, 
The analysis includes all submissions that exceed the baseline, as well as the baseline itself (6 models). We analyze examples of annotated opinion tuples that were not identified by all models under consideration, or tuples, which are absent in the annotated data but recognized by at least three models.

We found the following main causes of the differences between annotated tuples and the tuples extracted by the models.
First, the context of the sentence is sometimes not sufficient to reveal the sentiment or the holder of the opinion. For example, the sentence "By the way, Moscow occupies fourth place in the list of leaders for 2010" looks very positive towards Moscow but in fact leaders in road congestion (jams) are discussed, which means negative information about Moscow.

Second, in the task, we required participants to distinguish between the author's opinion and the opinion with the unknown holder. We thought that this is important for extracting the author's position. But this requirement was too difficult for models. In some cases, the short context of the sentence did not provide the necessary information. For example, from the sentence "Russians started the Dakar rally as favorites", it is difficult to understand if "favorites" is a general opinion or an author's opinion.

Next, we considered opinions between mention subjects that were not found by all the models.
In the following sentence,  all models did not find that the attitude of specialists towards UKRSPIRT is negative possibly because of the quite long distance between the mentions:
\begin{itemize}
\item "In addition, the very fact of the possibility of alternative supplies will undermine the [monopoly] of the state concern UKRSPIRT, which most often does not agree to increase prices for its products, specialists note." 
\end{itemize}

In the next sentence, the positive attitude from Syria to HAMAS, Hezbollah and Iran is presupposed in the sentence but was not revealed by all models: 

\begin{itemize}
\item In Tel Aviv, they believe that the price of the withdrawal of Israeli troops from the Golan Heights should be political concessions from Syria - weakening its ties with its strategic ally Iran and ending support for the Lebanese resistance movement Hezbollah and the Palestinian Hamas.
\end{itemize}

In the following sentence, the models did not reveal the negative attitude from V. But to the United States because of the long distance between entities and non-typical expression of sentiment:
\begin{itemize}
\item V. But suggested that "the court will not examine the factual objective side of the case, since the practice of hearing cases on charges of conspiracy in the United States is such that such charges automatically mean guilt."
\end{itemize}




\section{Conclusion}
In this paper, we presented the shared task of extracting structured opinions from Russian news texts. The task of the RuOpinionNE competition was to extract opinion tuples for a given sentence; the tuples are composed of a sentiment holder, its target, an expression, and the holder's sentiment towards the target. We prepared a new dataset of annotated opinion tuples. In total, the task received more than 100 submissions. The participants experimented mainly with large language models in zero-shot, few-shot and fine-tuning formats. The best result on the test set was obtained with fine-tuning of a large language model. We also compared 30 prompts and 11 open-source language models with 3-32 billion parameters in the 1-shot and 10-shot settings and found the best model and prompt.

\section*{Acknowledgements}

The study was conducted under the state assignment
of Lomonosov Moscow State University.

\bibliography{dialogue.bib}
\bibliographystyle{dialogue}

\appendix
\newpage

\section{Appendix A: Prompt for Baseline Method}
\label{sec:baseline}


\begin{table}[!htp]
    \centering
    \begin{tabular}{p{8cm}|p{8cm}}
        \hline
         Original Text of the prompt (Russian) & Translated version of the prompt (English) \\
         \hline
\selectlanguage{russian}
\footnotesize
Твоя задача состоит в том, чтобы проанализировать текст и извлечь из него выражения мнений, представленные в виде кортежа мнений, состоящих из 4 основных составляющих:

1. Источник мнения: автор, именованная сущность текста (подстрока исходного текста), либо 
\selectlanguage{british}  NULL. Key = Source; \selectlanguage{russian}

2. Объект мнения: именованная сущность в тексте (подстрока исходного текста).  \selectlanguage{british} Key = Target; \selectlanguage{russian}

3. Тональность: положительная/негативная \selectlanguage{british} (POS/NEG). Key = Polarity;\selectlanguage{russian}

4. Языковое выражение: аргумент, на основании которого принята результирующая тональность (одна или несколько подстрок исходного текста). \selectlanguage{british}Key = Expression;\selectlanguage{russian}

Если источник мнения отсутствует, то \selectlanguage{british}Source = NULL\selectlanguage{russian}. 
Если источником мнения является автор, то 
\selectlanguage{british}Source = AUTHOR\selectlanguage{russian}. 
В прочих случаях поле Source должно полностью совпадать с подстрокой исходного текста. Поля \selectlanguage{british}Target, Expression\selectlanguage{russian} всегда совпадают с подстроками текста.

Ответ необходимо представить в виде \selectlanguage{british}json\selectlanguage{russian} списка, каждый элемент которого является кортежем мнений. Каждый кортеж мнений это словарь, состоящий из четырех значений: \selectlanguage{british}Source, Target, Polarity, Expression.\selectlanguage{russian}

Для извлечённых \selectlanguage{british}Source, Target, Polarity, Expression\selectlanguage{russian} должно быть справедливо утверждение: ""На основании выражения \selectlanguage{british}Expression\selectlanguage{russian} можно сказать, что \selectlanguage{british}Source\selectlanguage{russian} имеет \selectlanguage{british}Polarity\selectlanguage{russian} отношение к \selectlanguage{british}Target\selectlanguage{russian}"".

Ниже представлены примеры выполнения задачи:

***Текст***

Премьер-министр Молдовы осудил террориста за бесчловечные и жестокие действия.

\selectlanguage{british}Source:\selectlanguage{russian} Премьер-министр Молдовы, \selectlanguage{british}Target:\selectlanguage{russian} террориста, \selectlanguage{british}Polarity: NEG, Expression:\selectlanguage{russian} бесчловечные и жестокие действия

***Текст***

Знаменитая актриса продемонстрировала человечность и простоту, достойную уважения публики.

***Ответ***
\selectlanguage{british}Source: AUTHOR, Target:\selectlanguage{russian} актриса, \selectlanguage{british}Polarity: POS, Expression:\selectlanguage{russian} продемонстрировала человечность и простоту, достойную уважения публики

Проанализируй таким же образом следующий текст.

***Текст***

\{text\}
         
\selectlanguage{british}
         &  
         \small
Your task is to analyze the text and extract from it expressions of opinion, presented as a tuple of opinions consisting of 4 main components:

1. Opinion source: the author, a named entity in the text (a substring of the source text), or NULL. Key = Source;

2. Opinion object: a named entity in the text (a substring of the source text). Key = Target;

3. Sentiment: positive/negative (POS/NEG). Key = Polarity;

4. Language expression: the argument on the basis of which the resulting sentiment is accepted (one or more substrings of the source text). Key = Expression;

If the opinion source is missing, then Source = NULL. 

If the opinion source is the author, then Source = AUTHOR. In other cases, the Source field must completely match the substring of the source text. The Target, Expression fields always match the substrings text.

The answer must be presented as a json list, each element of which is a tuple of opinions. Each tuple of opinions is a dictionary consisting of four values: Source, Target, Polarity, Expression.

For the extracted Source, Target, Polarity, Expression, the statement ""Based on the Expression, it can be said that Source has a Polarity relation to Target"" must be true.

Below are examples of completing the task:

***Text***

The Prime Minister of Moldova condemned the terrorist for inhumane and cruel actions.

Source: Prime Minister of Moldova, Target: terrorist, Polarity: NEG, Expression: inhumane and cruel actions

***Text***

The famous actress demonstrated humanity and simplicity worthy of public respect.

***Answer***

Source: AUTHOR, Target: actress, Polarity: POS, Expression: demonstrated humanity and simplicity worthy of public respect

Analyze the following text in the same way.

***Text***

\{text\}

         \\
         \hline
    \end{tabular}
    \caption{Textual prompt of the baseline submission; \{text\} refers to the input parameter of the sentence}
    \label{tab:my_label}
\end{table}

\end{document}